\documentclass[10pt,twocolumn,letterpaper]{article}

\usepackage[pagenumbers]{cvpr} 

\usepackage{graphicx}
\usepackage{amsmath}
\usepackage{amssymb}
\usepackage{booktabs}
\usepackage{algorithm2e}
\usepackage[numbers,sort&compress]{natbib}

\usepackage[pagebackref,breaklinks,colorlinks,citecolor=cvprblue]{hyperref}

\usepackage[capitalize]{cleveref}
\crefname{section}{Sec.}{Secs.}
\Crefname{section}{Section}{Sections}
\Crefname{table}{Table}{Tables}
\crefname{table}{Tab.}{Tabs.}

\usepackage[dvipsnames]{xcolor}
\usepackage{subcaption}

\definecolor{cvprblue}{rgb}{0.21,0.49,0.74}

\Crefname{ALC@unique}{Line}{Lines}

\newcommand{\rex}{\textsc{r}e\textsc{x}\xspace}
\newcommand{\gradcam}{\textsc{g}rad-\textsc{cam}\xspace}
\newcommand{\lime}{\textsc{lime}\xspace}
\newcommand{\shap}{\textsc{shap}\xspace}
\newcommand{\rise}{\textsc{rise}\xspace}
\newcommand{\xai}{XAI\xspace}
\newcommand{\pdc}{\emph{PDC}\xspace}
\newcommand{\dc}{\emph{DC}\xspace}
\newcommand{\hpe}{\emph{HPE}\xspace}

\def\paperID{3777} 
\def\confName{CVPR}
\def\confYear{2024}

\begin{document}

\title{MRxaI: Black-Box Explainability for Image Classifiers\\ in a Medical Setting}

\author{Nathan Blake, Hana Chockler, David A. Kelly,\\Santiago Calder\'{o}n Pe\~{n}a, Akchunya Chanchal\\
Department of Informatics,\\
King's College London\\
{\tt\small nathan.blake, hana.chockler, david.a.kelly, }\\{\tt\small santiago.calderon, akchunya.chanchal AT kcl.ac.uk}
}
\maketitle

\begin{abstract}
Existing tools for explaining the output of image classifiers
can be divided into white-box, which rely on access to the model internals, and black-box, agnostic to the model. 
As the usage of AI in the medical domain grows, so too does the usage of explainability tools. Existing work on
medical image explanations focuses on white-box tools, such as
\gradcam. However, there are clear advantages to switching to a black-box tool,
including the ability to use it with any classifier and the wide selection
of black-box tools available. On standard images, black-box tools are as precise
as white-box. In this paper we compare the performance of several black-box methods against \gradcam on a brain cancer MRI dataset. We demonstrate that most black-box tools
are not suitable for explaining medical image classifications and present a
detailed analysis of the reasons for their shortcomings. We also show that one black-box tool, a causal explainability-based \rex, performs as well as \gradcam.

\end{abstract}

\section{Introduction}\label{sec:intro}

Machine learning (ML) for classification in the medical domain is facing what has been described as a ``reproducibility crisis''. It is characterized by a model's published performance failing to generalize to the clinical setting, even when deployed in ostensibly the same setting ~\cite{hutson2018artificial, mcdermott2021reproducibility, volovici2022steps}. 
Hence, reported state-of-the-art performance is often not reproducible. For instance, a recent systematic review of ML applied to diagnosing COVID with radiographs found that out of $62$ studies, none were clinically relevant \cite{roberts2021common}. 
Even in more established diagnostic pathways, such as neuroimaging for intracranial hemorrhage, very few clinically relevant models exist out of thousands of developed models \cite{agarwal2023systematic}.

There are several factors contributing to this phenomenon. Exceptional performance is often achieved on highly curated benchmark datasets. While these have utility for model development, they are rarely representative of data from real clinical environments \cite{varoquaux2022machine}. 
Such benchmark datasets are often small, relative to the large datasets used in deep learning (DL), typically of the order of hundreds of patients. This is compounded by the fact that medical data is complex and hierarchically structured. 
For instance, MRI images taken with a single MRI machine may learn to distinguish instrumental rather than clinical features \cite{rajpurkar2023current}. Even when trained upon numerous MRI machines, the imaging protocol can influence downstream learning \cite{dewey2019deepharmony}. Even racial structures in datasets have been observed to contribute to a lack of generalizability in various disease classification tasks \cite{rajpurkar2022ai}.

If such factors are not carefully considered, the resultant models fail to learn the relevant features and do not generalize. This 'dataset' or 'distributional' shift can result in models with high published evaluation metrics which fail to apply to the clinical setting. To counter this potential misuse of ML, the use of counterfactual reasoning to distinguish correlation from causation is growing in popularity \cite{volovici2022steps}. 

Interest in explainability tools for medical applications has rapidly grown recently \cite{chaddad2023survey} as clinicians seek trustworthy models to assist with various medical tasks, notably including imaging tasks.
To this end, there are calls for explainable AI (XAI) to check the fidelity of medical classifiers \cite{rajpurkar2022ai, rajpurkar2023current}. By highlighting which features a classifier has learnt to utilize, the XAI algorithm could assess whether the classifier has been over-fit to learn spurious features. This would provide a post-processing quality control step to check that the classifier is indeed highlighting clinically relevant features. 

We assessed several popular XAI tools applied to a model developed to detect the presence of brain tumor in MRI images. This is the first paper to explicitly compare a suite of black-box XAI tools in a strictly medical imaging context, comparing performance to a white-box standard, \gradcam.

Detailed experimental results, including the datasets and evaluation, are submitted as a part of the supplementary material.
\section{Background}\label{sec:xai}

The most commonly used XAI tools in medical diagnostic imaging are Gradient-weighted Class Activation Mapping (\gradcam)~\cite{CAM}, Local Interpretable Model-agnostic Explanations (\lime)~\cite{lime} and SHapley Additive exPlanations (\shap)~\cite{lundberg2017unified}. Randomized Input Sampling for Explanation (\rise) is less popular in the medical literature but has attractive features, discussed below, making it worth considering.

\gradcam works in CNNs by using the gradients of a target class (e.g. a disease type) flowing into the final convolutional layer to produce a salience map of important regions. \gradcam is the most commonly used XAI tool for medical imaging applications \cite{gulum2021review, van2022explainable, chaddad2023survey}. However, it has been shown to be fragile in the sense that two imperceptibly different images with the same label can give very different explanations \cite{ghorbani2019interpretation}. It has also been found unsuitable for medical applications using certain DL architectures. For instance, in a brain MRI task the final convolutional layer of a DenseNet model was too small ($3\times3\times3$) to be clinically useful when upscaled to the full image size ($120\times120\times120$) \cite{wood2022deep}. 

A large disadvantage of any white-box method is the necessity of having access to the internals of the model. This may not be available depending upon the nature of various healthcare and commercial collaborations, and even between purely academic teams there may be a reluctance to share. This would only exacerbate the reproducibility crisis, where the explanations for a given model are not amenable to verification. Thus black-box XAI tools have a critical role in the verification of medical models. The following XAI tools are all model-agnostic.  

\lime, in the context of images, works by perturbing images from which predictions are obtained. These are then used to train a simpler model (often linear), from which the feature weights are extracted to build a map of their importance. The default perturbation used in \lime is to replace a pixel with the mean segment pixel intensity. Several medical researchers have used \lime for explaining DL models \cite{lime}.


\shap uses Shapley values, from game theory, to determine the marginal contribution of every feature (superpixels) to the model output \cite{lundberg2017unified}. Although generally popular, it is less commonly used for medical image explanations, perhaps due to its computational cost \cite{van2022explainable}. To calculate the exact Shapley values would require $2^F$ mutants, where $F$ is the total number of pixels. To find solutions in polynomial time, it leverages Owen values, trading accuracy for speed.

\rise computes the importance of each pixel by applying binary masks over an image \cite{petsiuk2018rise}. These masks are then upsampled using bi-linear interpolation, which seeks to avoid sharp edges for a smooth saliency map. During interpolation these upsampled masks do not remain binary but take values in $[0,1]$. The resulting mask is applied to the model to obtain a confidence score. Thus, the importance of each pixel is quantified based on the expected confidence score over all possible masks in which the pixel is observed.

The XAI tools described above, and indeed many similar tools, occlude, or otherwise perturb, various pixels in an input image to determine the difference such perturbations make on its classification. 
The nature of the perturbations might include setting pixel values to zero or the mean pixel value, with or without blurring at the edges. 
The strategy by which perturbed pixels are selected may impact explanations, and can, for instance, manifest in particularly rectilinear, and quite unnatural, explanations. 
Some of these will be impacted by the selection of tool parameters while others will be implicit in the nature of the XAI tool itself. 
These considerations are likely exacerbated by the relative homogeneity of medical images compared to normal image classification: the difference between a sub-species of dogs is larger than between sub-types of cancer, and so the latter may be more sensitive to the nature of XAI tool perturbations (and their associated parameters). Most XAI tools have a number of parameters, whether implicit or explicitly declared. 
Though default settings exist, these are rarely optimised for use in medical images, and their selection is non-trivial, especially for clinical end-users. Hence it is desirable for an XAI tool to have few, if any, parameters which require fine-tuning, or to default those parameters to domain specific values~\cite{van2022explainable}. 

A desirable trait for an XAI tool is to move beyond correlation and provide explanations that incorporate causal relationships~\cite{van2023explainable}. Tools, which account for such relationships, allow insights into potential biases which can then be removed \cite{van2019eliminating}. 
Causal \underline{R}esponsibility \underline{eX}planations (\rex) \cite{chockler2021explanations} is an XAI tool 
based on causal reasoning~\cite{Hal48}. 

\rex constructs an approximation of a causal explanation~\cite{HP05,Hal48} by first ranking the pixels of a given image
according to their importance for the classification, and then constructing an explanation greedily from the saliency landscape of the pixels. The ranking is an approximate degree of responsibility~\cite{CH04}, 
computed on the coarsely partitioned image, and updated by refining
the areas of the input with high responsibility, until a sufficiently fine partition is reached. 
To compute the degree of responsibility, \rex generates mutants (images with some parts masked) and queries the classifier on these mutants. If a non-masked area is sufficient to get the same classification as the original
image, it is a cause for the classification, with the degree of responsibility of each part based on the number of parts in the area.

\section{Models, Methods and Measures}\label{sec:methods}
We investigated the performance of several image-based XAI tools on low diversity medical data. XAI tools are frequently trained on standard datasets, such as ImageNet~\cite{imagenet}. These images exhibit a relatively high degree of diversity, so that a dog, for example, may appear against many different backgrounds, in different sections of the image. MRI images almost completely lack this diversity, with brains at consistent locations, sizes and pixel densities. Occlusion techniques and defaults developed for models trained on diverse data may be inappropriate for models trained on low diversity data. The model may be overly sensitive to occlusions, or fail entirely. 

\subsection{Model and dataset}\label{subsec:model}

The model is a pre-trained CNN based on the ResNet 50 architecture \cite{legastelois2023challenges}. Brain magnetic resonance imaging (MRI) data was obtained from The Cancer Imaging Archive, as published by Buda et al. \cite{buda2019association}, 
and made publicly available on kaggle~\cite{kaggle}. This curated dataset of $110$ pre-operative patients with low grade gliomas (LGG) was gathered from five US institutions. All of the images are axial slices of the brain and include fluid-attenuated-inversion recovery (FLAIR) sequence, while $101$ also have pre-contrast sequence and $104$ have post-contrast sequence patient images. Each patient had between $20$ and $88$ slices taken, with a total of $3,929$ images. Each image contains $3$ channels, one for each of the three sequences. All images are  $256\times256\times3$. A radiologist annotate the FLAIR images into binary masks of ``tumor'' or ``no tumor''. This provides a measure of the ground truth, or Human-Provided Explanation (\hpe), by which to quantify the performance of the XAI tools.

As there is no clear definition of an explanation of a negative classification, we have retained only the positive (diseased) images for assessment, leaving $1370$ images. Of these $1370$, $118$, $\approx9\%$ of the data, are false negatives, where the model reports no tumor, even though the \hpe indicates tumor. As the goal is to explain positive classifications, we also exclude the false negatives. The failure modes of each tool are interesting in and of themselves and merit further study.


This is not a clinically motivated task, but one designed to establish the validity of XAI tools in a medical imaging context, based on the assumption that diagnostically significant features coincide with the \hpe. This may not be strictly true, as, for instance, tumors present in a brain may distort the ventricles distal to a tumor and so even though a tumor may not be present in a given MRI slice, clinical features may still be present. However, as the CNN was trained on MRI slices as opposed to whole 3D images, the classifier was forced to ignore such features and focus on what distinguishes the slices: the increased signal intensity of pixels containing tumor.

\subsection{Quantifying explainability performance}

Very few studies of XAI tools in the medical domain have been subject to clinical scrutiny \cite{gulum2021review}. In part this is due to success criteria for explanations being more subjective than standard measures of classifier performance such as ROC analysis \cite{van2022explainable}. Ultimately, the 'best' measure of XAI performance is task specific. One method is to allow clinicians to assess XAI outputs, judging their usefulness on whatever criteria they deem most appropriate. However, this is not without its own limitations, as humans are prone to biases, such as using XAI to confirm existing beliefs and decreased vigilance of AI systems \cite{ghassemi2021false}. In addition, this method requires copious amounts of clinician input, which is a scarce resource. Therefore we utilise a clinically annotated  segmentation (\hpe) as a baseline for XAI performance. 

All XAI tools tested provide explanations in the sense that the pixels indicated by the tool are sufficient to generate a positive classification by the model. It is categorically not the case, however, that each explanation is equally good. We assess tool output against human expectation via comparison with the \hpe. Our measure, detailed in~\cref{subsec:pdc}, allows us to grade tool explanations against the \hpe and therefore provide a ranking of tools by human explanation fidelity. The underlying assumption of this procedure is that a better explanation is one that better coincides with an explanation provided by a clinician. 

\subsubsection{Penalised Dice Coefficient (PDC)}\label{subsec:pdc}

Even with a clinical \hpe against which to compare XAI outputs, there are many ways to construct a measure of performance. The 
S\o{}rensen–Dice coefficient (\dc) is one such statistic, used to gauge the similarity of any two samples~\cite{dic45,sor48}. 
It is a common metric to assess the similarity of medical images in segmentation tasks, where one image is the result of a segmentation model and the other is a `ground-truth' mask, such as a \hpe. 
Given two sets of pixels $X$ and $Y$, we have

\begin{equation}
    DC = \frac{2 \mid X \cap Y \mid}{\mid X \mid + \mid Y \mid.}\label{eq:dc}
\end{equation} 

Unfortunately \dc does not account for the distance between an explanation and the \hpe, nor the size or the number of explanation segments. These are important as the farther an explanation is from the \hpe, the less clinically useful it is. Similarly, an explanation that is too large can distract attention, and one too small
can lead to missing clinical information. An explanation erroneously consisting of multiple non-contiguous areas is also distracting. We therefore define a new measure, the \emph{Penalized Dice Coefficient (\pdc)}. It captures the difference between the \hpe and a given explanation, $exp$, taking the comparative areas, the distance between the areas, and the number of areas as follows.
%
%
The function $E$ is the standard euclidean distance, measured between the centers of the \hpe and $exp$, normalized against the maximum possible distance from the center of the \hpe ($E_{max}$), captured
as $d = 1 - E/E_{max}$; $d=0$ means that the centers are in the same location, and $d=1$ that the centers are maximally far from each other.



We also calculate the ratio between the \hpe area and $exp$ area. As we require the \pdc to always be in the range $(0, 1)$ for easy comparison, we define the ratio $r$ as:

\begin{equation}
   r = 
\begin{cases} 
s \frac{exp_{\text{size}}}{\hpe_{\text{size}}} & \text{if } exp_{\text{size}} < \hpe_{\text{size}} \\
b \frac{\hpe_{\text{size}}}{exp_{\text{size}}} & \text{if } \hpe_{\text{size}} < exp_{\text{size}} \\
1 & \text{if } exp_{\text{size}} = \hpe_{\text{size}}
\end{cases} 
\end{equation}

where $s$ and $b$ are parameters, between $(0, 1)$, which allow users to define how much they wish to penalise against explanations that are too small or too big, respectively. The ratio $r$ is thus bounded in the range $(0, 1)$. 

We combine the distance and area ratio with the \dc (the latter ensuring that explanations which overlap with the \hpe score higher than non-overlapping explanations) to form the summary statistic of the \pdc:

\begin{equation}
\pdc = \frac{d + r + DC}{3}
\end{equation}

As each component $d$, $r$ and \dc is in $[0, 1]$, the \pdc is also in this range. 1 indicates a perfect alignment of the \hpe and explanation area and 0 indicates an empty explanation. Unlike the \dc, which simply returns $0$ if two masks do not overlap to any degree, 
the \pdc is always a strictly positive number, unless no explanation is provided at all, but approaches $0$ as the different penalties for size and location come into effect.

In many cases XAI tools give an explanation with multiple non-contiguous areas. In this case the \pdc for a single explanation mask is given as the mean of all non-contiguous areas. We also report the number of non-contiguous areas found in the explanations. 


                

\subsection{Comparing tools output}
Unfortunately, the different \xai{} tools do not produce outputs which are directly comparable or immediately amenable to the \dc or \pdc. Both \lime and \rex provide boolean-valued masks in addition to heatmaps. Comparing these masks is much easier than comparing heatmaps. We use these masks directly when comparing against \hpe. 

\shap, \rise and \gradcam do not provide these masks directly, so we need to extract them from the heatmaps. We do this via a method similar to that used by \rex to produce its minimal passing explanations. We choose those pixels which are most highly ranking in the heatmap and query the model on these pixels only. If this satisfies the classification requirement, we quit there, otherwise we add in the next set of pixels and so on until we have the necessary classification. This gives us a direct assessment over the quality of the heatmap. 

We then extract the \dc, \pdc and number of non-contiguous areas (count) of every positively labelled image, using each XAI tool.

\gradcam is unique amongst the XAI tools under consideration in this paper in that it is the only white-box method. Having direct access to the model weights should theoretically confer it a significant advantage, given that the setting is not adversarial. Hence, it serves as the gold-standard by which to compare the black-box XAI tools.

\section{Results}\label{sec:results}
We evaluated the XAI tools against the \hpe (Human-Provided Explanation) using our \pdc. We also inspected tool performance in order to understand why they succeed or fail. None of the XAI tools have been optimized for this particular task. Such an optimization for every tool would be impractical. Moreover, parameter fine-tuning suggests prior knowledge on the part of the user as to appropriate values. We wished to determine the effectiveness of a tool without such knowledge.
We proceeded with the most parsimonious case of using each tool's default parameters, except for the number of mutants generated which was set at 2000. In the case of \rex, there is no default distribution to decide partitioning, so we default to uniform as the least informative. 

\begin{figure*}[htbp]
  \centering
  \includegraphics[width=0.45\textwidth]{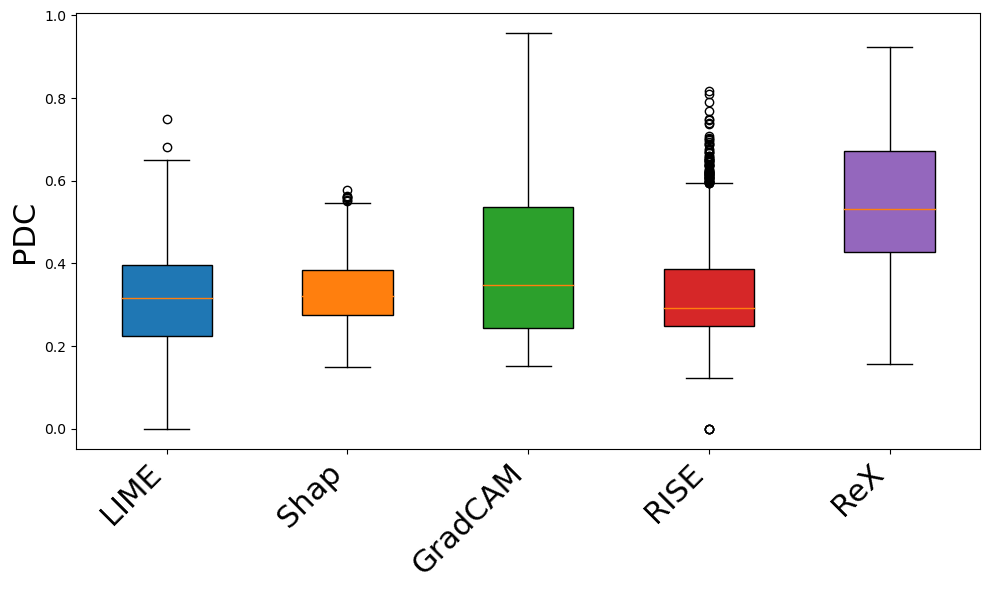}
  \hfill
  \includegraphics[width=0.45\textwidth]{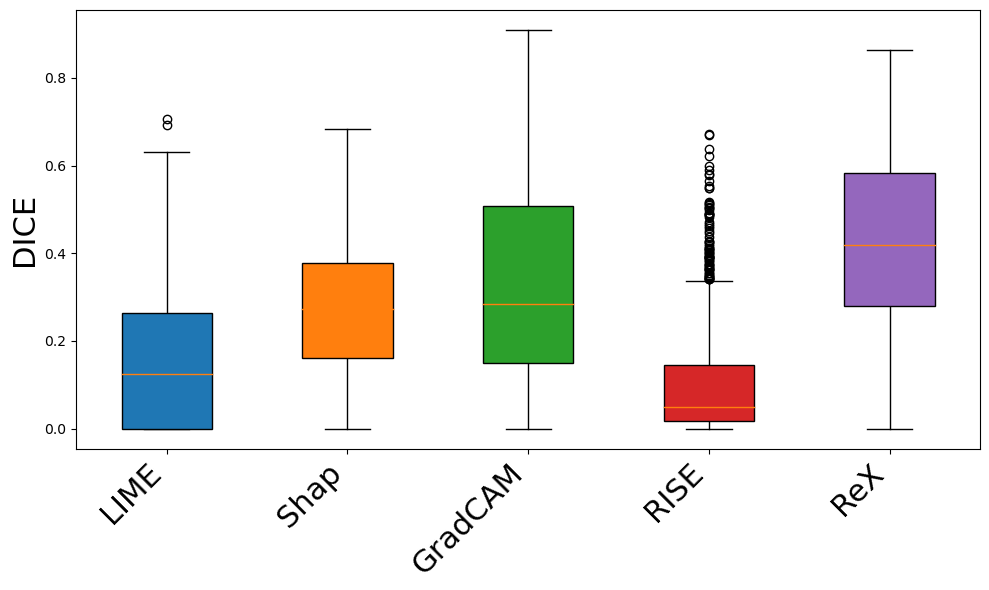}
  \caption{Box and whisker plots for the XAI tools as assessed by \pdc (left panel) and $DC$ (right). Each box delineates the interquartile range (IQR), with the top and bottom edges representing the 75\textsuperscript{th} (Q3) and 25\textsuperscript{th} (Q1) percentiles, respectively. The horizontal line inside the box indicates the median. Whiskers extend to the most extreme data point which is no more than 1.5 times the IQR away from the box. Points outside the whiskers are considered outliers and are represented as circles.}
  \label{fig:PDC_DICE_plot}
\end{figure*}

\begin{figure}[h] 
  \centering
  \includegraphics[width=1\linewidth]{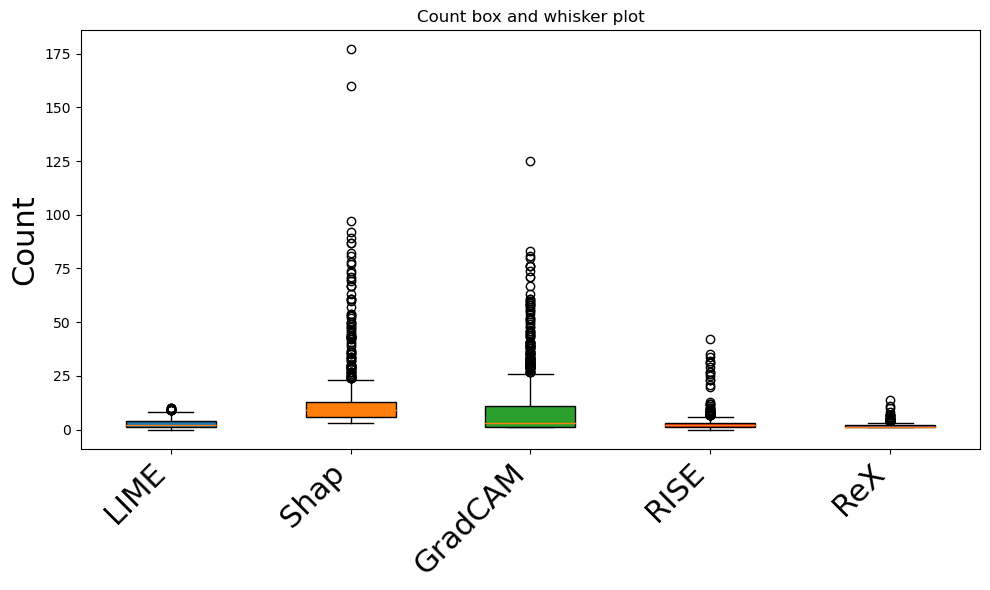}
  \caption{Box and whisker plot of the count of non-contiguous regions given by each XAI tool}
  \label{fig:Count_plot}
\end{figure}

Given size penalization parameters $b = s = 1.0$ (not additionally penalizing explanations that are too large or too small) we present the relative performance of each tool in~\cref{tab:results} and ~\cref{fig:PDC_DICE_plot} and~\ref{fig:Count_plot}.

\begin{table*}[]
    \centering
    \begin{tabular}{c||r|r|r}
    Tool & Count & \dc & \pdc \\ \toprule
    \lime   & 2.41 ± 1.95 & 0.16 ± 0.16 & 0.28 ± 0.17  \\
    \rise   & 2.64 ± 3.43 & 0.10 ± 0.12 & 0.33 ± 0.13  \\
    \shap   & 11.96 ± 13.21 & 0.27 ± 0.14 & 0.33 ± 0.08  \\
    \gradcam   & 9.08 ± 13.37 & 0.33 ± 0.22 & 0.41 ± 0.2  \\
    \rex   & \textbf{1.45 ± 1.01} & \textbf{0.42 ± 0.2} & \textbf{0.55 ± 0.16}  \\
    \end{tabular}
    \caption{Mean performance ± 1 standard deviation for all five XAI tools assessed by three measures. Count refers to the number of non-contiguous areas returned by a tool. Best performance per measure in \textbf{bold}. The best value for all parameters would be $1$.}
    \label{tab:results}
\end{table*}

Across all three measures of performance, \rex is the best performing tool, exceeding even \gradcam, which is generally the next best performing tool. The remaining XAI tools have comparable performance as measured by \pdc, though differences manifest in the $DC$ and count, with \rise performing particularly poorly as measured by $DC$ and \shap as measured by count.

\begin{figure*}[!tb] 
\centering
\includegraphics[width=0.97\textwidth]{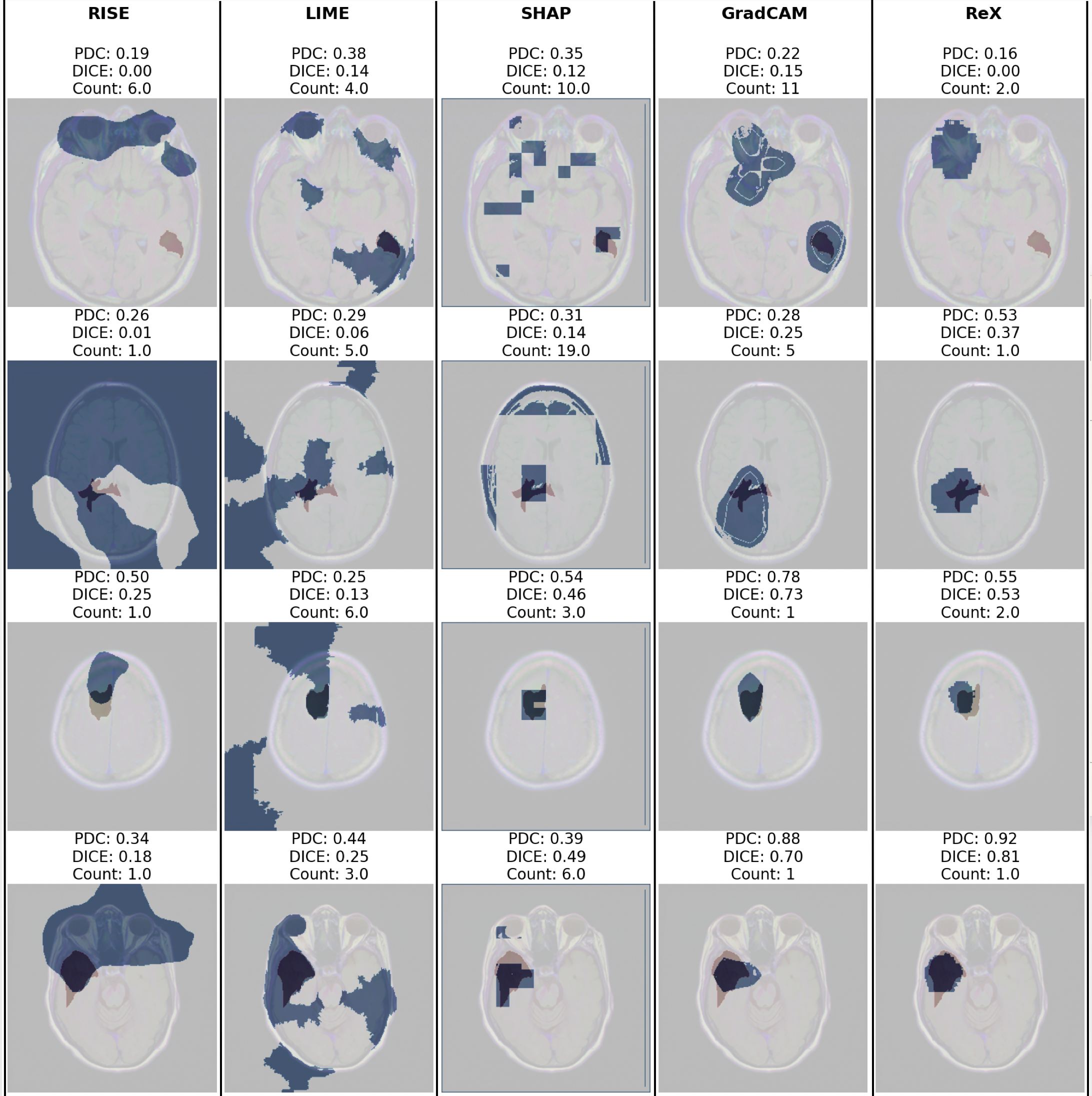}
\caption{A selection of XAI outputs for images representing, from top row to bottom row, the \textbf{worst, median, mean,} and \textbf{best} performing \rex results as measured by \pdc. All tools give differently formatted outputs, these have been homogenized for ease of comparison. Gray background shows the underlying MRI, pink patches show the tumor as defined by \hpe and the blue regions are the tools' respective explanations.}
\label{Results_Panel}
\end{figure*}

For the purposes of visualization we show four rows of results corresponding to the worst, median, mean and best \rex \pdc respectively in ~\cref{Results_Panel}.


\subsection{Tool analysis}\label{subsec:tools}

In this section we inspect the explanations provided by the XAI tools, and investigate their performance and their failures modes. 

\begin{figure*}[htbp]
    \centering
    \begin{subfigure}[t]{0.185\textwidth}
        \centering
        \includegraphics[scale=0.33]{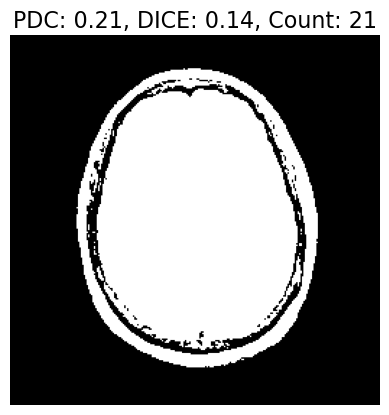}
        \caption{\gradcam}\label{subfig:gradcam}
    \end{subfigure}%
    ~ 
    \begin{subfigure}[t]{0.185\textwidth}
        \centering
        \includegraphics[scale=0.33]{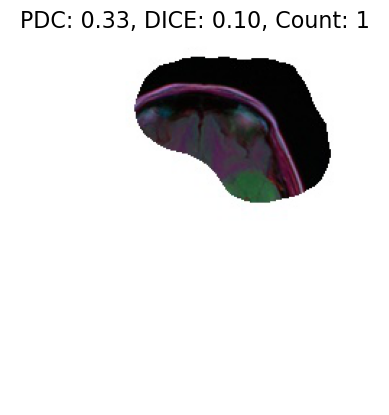}
        \caption{\rise}\label{subfig:rise}
    \end{subfigure}
    ~ 
    \begin{subfigure}[t]{0.185\textwidth}
        \centering
        \includegraphics[scale=0.61]{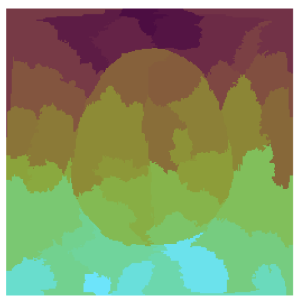}
        \caption{\lime}\label{subfig:lime}
    \end{subfigure}
    ~ 
    \begin{subfigure}[t]{0.185\textwidth}
        \centering
        \includegraphics[scale=0.33]{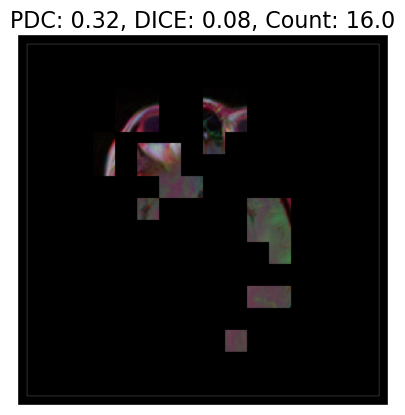}
        \caption{\shap}\label{subfig:shap}
    \end{subfigure}
    ~ 
    \begin{subfigure}[t]{0.185\textwidth}
        \centering
        \includegraphics[scale=0.33]{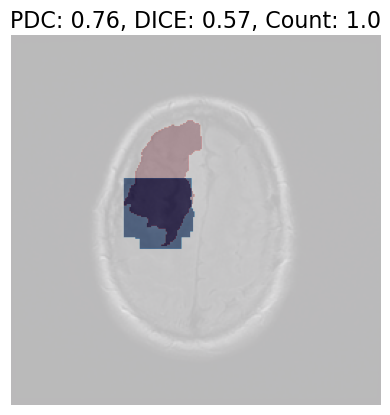}
        \caption{\rex}\label{subfig:rex}
    \end{subfigure}
    \caption{A selection of strange results from XAI tools (not on the same image). \ref{subfig:gradcam} shows the \gradcam binary mask. 
    Note the separation of brain and skull and the islands contained therein. \ref{subfig:rise} shows the output of \rise. \rise often localizes 
    the front of the skull, even when eyes are not present. \ref{subfig:lime} shows the segmentation map produced by \lime 
    (not \lime's final output). As this map shows, \lime often produces anatomically meaningless segments. 
    \ref{subfig:shap} shows the \shap binary map. As well as the tumor, numerous other 
    regions of the brain are also erroneously given as parts of the explanation. \ref{subfig:rex} shows \rex hybrid mask. 
    Note the sharp line of the explanation that has no anatomical justification and cuts out the remainder of the tumor.}\label{fig:XAI_tools}
\end{figure*}

\textbf{\gradcam} is the only white-box XAI tool we investigated, \gradcam is the second best performing tool except by count, in which it performs poorly. This divergence is explained by the fact that although \gradcam can localise well to the tumor, it also often includes unassociated regions of the head (though it seldom shows areas outside of the head, which helps imbue trust with clinicians), as demonstrated in ~\cref{subfig:gradcam}. These partitionings tend to separate the brain and the skull (called the subarachnoid space), sometimes containing multiple small 'islands', which increases the count of non-contiguous areas. This separation is perhaps related to a similar phenomenon observed in brain imaging of explanations localising to eyeballs -- the sharp gradients of such anatomical features confuse the model as sharp gradients are also a common feature of tumors.

From a qualitative perspective, even though \gradcam has a tendency to return multiple areas of explanation, these areas are often contained within a region. This would draw a clinicians eye to a single region (see top row of ~\cref{Results_Panel}), thus the \gradcam count score is not as bad as first appears. 

As \gradcam does not iteratively query the model, unlike black box XAI tools, it is comparatively fast. Speed has been cited as a critical for clinical applications \cite{chaddad2023survey}. However, the necessity for speed varies greatly by the nature of the clinical task. In intra-operative applications every second waiting for results could extend the length of surgery. For screening and triaging tasks, results could be expected to take hours, days and even weeks. Hence, slower methods could have utility, especially if they demonstrate superiority.  

\textbf{\rise} creates random rectangular masks which it overlays on an image. The number and size are controlled by parameters. We generate $2,000$ mutants, in keeping with the average amount of work performed by \rex. We set the other parameters to their default values. 
It is the least complex of the black-box tools we investigated, and is the median performer in terms of \pdc and count. 
However, the \dc values for \rise are the worst. 
The tool has a tendency of returning explanations at the front of the brain or explanations that fill the majority of the image (~\cref{subfig:rise}). 
To examine this, we calculated the average dice coefficient of the random masks produced by \rise against a mask containing the front region of the head. As we used the same seed for all experiments, the mask production is the same for all images. The average \dc of $0.002$ indicates that occlusions almost never covered these areas, hence their over-representation in passing mutants.

There were no duplicates in the mutants produced by \rise with our chosen seed.
The area a mutant covers with our parameters is approximately $10\%$ of the image. As the model is overly generous in what it accepts, the mutants that \rise produces do not cover enough of the image to force failing cases. Passing cases constitute $77\%$ of the mutants on our dataset. It has been long known in the testing community that a balance of passing and failing cases in required for a good-quality output \cite{abreu2007accuracy} and it is also the reason for outputting wrong regions of the image as the explanation here.
The result is a heatmap which is too hot in general. The concentration of explanations towards the front of the head is due to the particular set of mutants generated. This would likely change given a different random seed.

\textbf{\lime} is unique amongst the XAI tools we tested in that it uses a segmentation algorithm, rather than rectilinear partitioning, to generate occluded mutants (\cref{subfig:lime}). The resulting segments are generally larger than pertinent to neuro-anatomical features. Nor are they anatomically meaningful, though this could be achieved by segmentation algorithms dedicated to brain images~\cite{ghazi2022fast}. In our dataset, there was an average of $40$ segments per image. The probability of mutants sharing at least one segment approaches $1$ after just $19$ mutants. As \lime with default parameters generates $1,000$ mutants, the best case scenario is
that there exists a subset of $\approx25$ mutants that share a segment. In practice, this is likely to be much higher. 

Further, the default use of the mean segment value to generate mutants is not clinically meaningful. These result in mutants which are not diverse, likely a consequence of the relative homogeneity of MRI images compared to standard images for which \lime was developed. Additionally, as has been noted in the context of histopathological images \cite{graziani2021evaluation}, \lime is non-deterministic: given the same image and random seed it can produce different explanations. This is an undesirable trait, especially in a medical context in which consistency is paramount.

\textbf{\shap} is computationally expensive to run. It is the lowest performing in terms of count, on average returning $12$ non-contiguous regions. As seen in \cref{subfig:shap} these areas tend to be distant from each other and from the \hpe. This is ameliorated by the fact that amongst these many areas is the tumor, hence its \dc and \pdc are commensurate with the other tools. However, from a clinical perspective having so many disparate areas in an explanation is highly undesirable, as each area requires clinical scrutiny for confirmation.  

The computation of Shapley values requires considering all possible mutants to quantify the contribution of each pixel. To avoid this computational overhead, \shap relies on Owen values to estimate the contribution of a collection of pixels, which is a valid approximation of Shapley values \cite{Okhrati2021AValues}. As a consequence, the granularity of the explanations is constrained by the minimal grouping of pixels utilized.

Additionally, prior research has demonstrated that reliance on a single iteration can introduce bias into the explanations \cite{Chen2022ExplainingValues}. In the context of this experiment, this bias manifests in the form of certain images consistently yielding explanations that are invariably 0 throughout the whole image, where no mutant was able to induce a change in the image's classification. \shap requires the specification of a blur window in order to generate mutants of images. A small scale investigation suggested to us that there was no one ideal value for this setting. We utilized a $64 \times 64$ window in our final experiments, noting that different window sizes can produce different, including empty, results.

\textbf{\rex} is the best performing XAI tool by all measures. It is comparatively fast, in the order of seconds, due to its dynamic mutant generation which stops when no new mutants increase the measure of causal responsibility. 
Its explanations usually coincide with the tumor to at least some extent, thus producing a higher \dc and \pdc relative to the other tools. 
Additionally, it seldom returns spurious explanations with multiple areas, resulting in a low count.
However, as with other tools, it is not immune to the effect of the model treating eyeballs as signs of cancer.  

It is perhaps surprising that \rex outperforms \gradcam, but on this particular task it is highly likely that the model is suboptimal, despite a high test accuracy ($\approx91\%$). This is a very common feature of deep learning models in a medical context ~\cite{hutson2018artificial, mcdermott2021reproducibility, volovici2022steps}. As \gradcam is a white-box method, it is more sensitive to over-fitting accrued in the training process \cite{ghorbani2019interpretation},
and is likely to improve significantly on a thoroughly curated and validated dataset and model.

We occasionally see straight edges in \rex explanations (\cref{subfig:rex}). Straight edges are extremely unlikely in any natural phenomenon, including tumors. Their presence is explained by the effect of the spatial search algorithm. This places a square box centered over the region of highest responsibility. This square may grow, contract or move until an explanation is found. As this square is initially placed over the pixels of highest responsibility, this procedure does not, in general, take very long. However, in some cases, a box grabs enough of the tumor to satisfy the model, even though a better explanation (\ie one with a higher \pdc) may exist. Straight lines in the explanation may point to the presence of a better explanation in the direction of those straight lines.

\subsection{Limitations}\label{sec:limitations}

All of the XAI tools were used in their default settings, with the exception of mutant budget. This was set to 2000 where applicable. It is likely that each could be optimized for clinical applications to improve the performances reported here. However, it would be impractical to exhaustively search the entire parameter space for each tool. Even if possible, great care would be required to avoid over-fitting the tool parameters to perform well on a particular model and dataset, but being unable to generalize to new datasets. Therefore assessing each tool with its default settings provides the most parsimonious comparison.  

Although the \pdc has \textit{prima facie} validity in that it rewards explanations closer, and of a similar size, to the \hpe, it requires validation in the clinical setting. To this end we plan to correlate \pdc with clinicians' assessments to develop the first clinically validated measure of XAI tools. These results are based upon a single clinical dataset and would need to be repeated on several such datasets in order to confirm the generalisability of our findings.
\section{Related Work}\label{sec:relwork}

Much of the focus of XAI in the medical domain is on the need for it and its required characteristics, how it should be incorporated into medical workflows and how to evaluate its outputs. There are also many studies which employ XAI in a medical context. Far fewer publications have focused on directly comparing XAI tools as we have here. 

In one study comparing several white-box XAI tools for chest x-ray images, \gradcam was found to be the best performing, supporting our decision to use it as a gold-standard comparator \cite{saporta2022benchmarking}. This study also used \hpe to compare XAI tool outputs, but additionally had blinded clinicians again annotate the images in order to create a comparator. By doing so they found that \gradcam struggled with pathologies that were small or irregular in shape. We have also noted that all the XAI tools struggle to match the contours of irregular shaped tumors. The very occasional exception to this is \lime, which at times provided at least partial tracing of complex shapes. This is likely due to its segmentation step sometimes successfully delineating the complex contours of a tumour.

Arun \textit{et al.} similarly compared a number of white-box XAI tools and found that none performed as well as dedicated segmentation DNNs \cite{arun2021assessing}. They stress that the tools failed on a number of clinical benchmarks, highlighting the performance level required of clinically orientated XAI tools over and above normal imaging standards.

That different XAI tools will give different explanations with the same model and dataset has been noted in various medical contexts including; histopathology (CAM variants vs \lime variants) \cite{graziani2021evaluation}, blood test results (interpretable models vs \shap) \cite{onari2022comparing} and electronic healthcare records (\shap vs \lime) \cite{duell2021comparison}. We have similarly noted that the XAI tools often give conflicting information to one another. As they are evaluating the same model, this can only be an artefact of the XAI tools themselves.


Although several papers have compared various XAI tools, as far as we can discern this is the first paper comparing a suite of black box XAI tools against the presumed gold-standard \gradcam on MRI data.
\section{Conclusions}\label{sec:conclusion}

The XAI tools considered in this paper all have their limitations when applied to tumor detection in MRIs. \rex{} performs the best across all measures of performance, outperforming the gold-standard \gradcam which is significant given that the latter is a white box method. 

If XAI tools are to find a place in nascent clinical AI workflows, then their performance must be demonstrable to a degree commensurate to the task. \rex{} is under active development and assessment.  Discussions with consultant neuroradiologists and neuroimaging professors will guide the tool's development in the clinical setting. Future work will include assessing the tool with clinically validated deep learning models, trained on large, multi-site datasets with annotations performed by leading cancer imaging experts. 

In addition to this, future work will also focus upon including a segmentation process for the partitioning of images, similar to \lime but tested upon medical images to ensure more clinically relevant partitions. The nature of the occlusions themselves will also be explored, as replacing image pixels with zero, or mean/median values is not clinically meaningful. It may be possible to use variational autoencoders to replace occluded areas with essentially simulated data allowing, for instance, tumorous regions to be replaced with healthy looking pixel data. This has been demonstrated on 2D electrocardiogram data \cite{van2022improving}.

Although \rex outperformed its competitors here, it is possible that the other XAI tools could be improved to make them more amenable to clinical tasks. We will explore several modifications that might make these general XAI tools more suitable to medical images.

\newpage
{\small
\bibliographystyle{ieeenatcvpr}
\bibliography{medex,all}
}

\end{document}